\documentclass{article}

\usepackage{PRIMEarxiv}

\usepackage[utf8]{inputenc} 
\usepackage[T1]{fontenc}    
\usepackage{hyperref}       
\usepackage{url}            
\usepackage{booktabs}       
\usepackage{amsfonts}       
\usepackage{nicefrac}       
\usepackage{microtype}      
\usepackage{fancyhdr}       
\usepackage{graphicx}       
\graphicspath{{media/}}     
\usepackage{color}
\usepackage{multirow}
\usepackage{tabularx}
\usepackage{array}
\usepackage[section]{placeins}
\usepackage{appendix}

\newcolumntype{C}{>{\centering\arraybackslash}X}

\title{FuXi-DA: A generalized deep Learning data assimilation framework for assimilating satellite observations}

\author{
  Xiaoze Xu\textsuperscript{1,2}\thanks{This work was completed while Xiaoze Xu was interning at SAIS.} \thanks{These authors contributed equally to this work: xuxiaoze1998@163.com, xiuyu.sxy@gmail.com}, 
  Xiuyu Sun\textsuperscript{2}\footnotemark[2], 
  Wei Han\textsuperscript{3}\thanks{Corresponding authors: hanwei@cma.gov.cn, lihao$\_$lh@fudan.edu.cn}, 
  Xiaohui Zhong\textsuperscript{4,2},
  Lei Chen\textsuperscript{2}, 
  Hao Li\textsuperscript{4,2}\footnotemark[3]
  \\
  \\
  \textsuperscript{1}School of Atmospheric Physics, Nanjing University of Information Science and Technology
  \\
  \textsuperscript{2}Shanghai Academy of Artificial Intelligence for Science (SAIS)
  \\
  \textsuperscript{3}CMA Earth System Modeling and Prediction Centre (CEMC), China Meteorological Administration
  \\
  \textsuperscript{4}Artificial Intelligence Innovation and Incubation Institute, Fudan University
}
\begin{document}
\maketitle

\begin{abstract}
Data assimilation (DA), as an indispensable component within contemporary Numerical Weather Prediction (NWP) systems, plays a crucial role in generating the analysis that significantly impacts forecast performance. Nevertheless, the development of an efficient DA system poses significant challenges, particularly in establishing intricate relationships between the background data and the vast amount of multi-source observation data within limited time windows in operational settings.
To address these challenges, researchers design complex pre-processing methods for each observation type, leveraging approximate modeling and the power of super-computing clusters to expedite solutions.
The emergence of deep learning (DL) models has been a game-changer, offering unified multi-modal modeling, enhanced nonlinear representation capabilities, and superior parallelization. These advantages have spurred efforts to integrate DL models into various domains of weather modeling. Remarkably, DL models have shown promise in matching, even surpassing, the forecast accuracy of leading operational NWP models worldwide. This success motivates the exploration of DL-based DA frameworks tailored for weather forecasting models.
In this study, we introduces Fuxi-DA, a generalized DL-based DA framework for assimilating satellite observations. By assimilating data from Advanced Geosynchronous Radiation Imager (AGRI) aboard Fengyun-4B, FuXi-DA consistently mitigates analysis errors and significantly improves forecast performance. Furthermore, through a series of single-observation experiments, Fuxi-DA has been validated against established atmospheric physics, demonstrating its consistency and reliability.
\end{abstract}


\section{Introduction}
Accurate weather forecasts are crucial for saving lives, emergency management, mitigating disaster impacts, and preventing economic losses due to severe weather events. 
At present, weather forecasts primarily rely on numerical weather prediction (NWP) models \cite{bauer2015quiet}. These traditional models generate weather forecasts by solving the governing partial differential equations based on the current state (called the analysis) of atmosphere and surface as initial conditions.
Since 2022, a new generation of deep learning (DL) models for medium-range weather forecasting has been recognized for their potential in medium-range weather forecasting, demonstrating forecast performance comparable to leading NWP models \cite{pathak2022fourcastnet,bi2023accurate,chen2023fuxi,lam2023learning}. 
Weather forecasting is essentially an "initial value" problem, where the greatest source of uncertainty is the initial conditions \cite{bjerknes1904problem}. 
Therefore, the accuracy of these initial conditions is of paramount importance, regardless of the forecasting method used \cite{Lorenz1963,lorenc1986analysis}.
Currently, both NWP and DL-based weather forecasting models depend on the analysis, which is derived from a combination of NWP output and observation through data assimilation (DA) systems, ensuring the highest possible accuracy of the initial conditions. The DA systems establish a foundation for reliable weather forecasting and are equally important as weather forecasting models.

Over the past few decades, advancements in NWP models and DA systems have significantly improved the accuracy of weather forecasts \cite{bauer2015quiet}.
Modern operational DA systems need to process a massive number of observations, including satellite data, radiosondes, aircraft and surface observations within a one-hour time window to produce the analysis for forecasting models \cite{barker2012weather}.
Despite these advances, the development of DA systems has faced several challenges.
One primary challenge is the significant volume of observations, many of which are compromised by their inadequate quality \cite{bauer2010direct,li2022satellite}. This issue, coupled with the extensive computational resources and time required for data processing, forces operational DA systems to find a balance between increasing the volume of assimilated observation and maintaining the timeliness of forecasts. 
For instance, while the European Centre for Medium-Range Weather Forecasts (ECMWF) receives 800 million observations daily, only 60 million of these pass the quality control for integration into the Integrated Forecasting System (IFS) \cite{ECMWF_observation}. Moreover, satellite observations are predominantly assimilated under clear-sky conditions, neglecting valuable cloud and precipitation data, resulting in the use of only 5–10\% of total satellite data in global weather forecasts \cite{bauer2015quiet,geer2018all}.
Remarkably, this small fraction contributes to the majority of relevant information for weather forecasts \cite{bauer2015quiet,eyre2022assimilation}.
Another challenge lies in the complexity of traditional DA workflows, which results in substantial delays from the acquisition of new observations to their operational application. The ever-increasing volume of available observations amplifies this challenge, posing greater scientific challenges for DA systems.
Notably, satellites measure variables such as radiances, brightness temperatures, or reflectivities, which do not directly correspond to model variables like temperature, humidity or ozone.
Consequently, observation operators must be developed to convert model variables into model equivalents of the observed data for comparison, a process that not only must be tailored for each new type of observation \cite{eyre2022assimilation}, but also introduces additional errors \cite{watts2004identification}.
Lastly, widely adopted DA methods, such as the four-dimensional variational (4D-Var) \cite{rabier1998extended} and ensemble-variational (En-Var) method \cite{bannister2017review} are computationally expensive.
These challenges underscore the need for continuous innovation in DA techniques to enhance the efficiency and accuracy of weather forecasting in the face of growing data volumes and computational demands.

In recent years, DL models have demonstrated remarkable potential across various aspects of weather and climate modelling, catalyzing numerous efforts to integrate DL models with DA techniques \cite{cheng2023machine}. 
These efforts have largely focused on conducting proof-of-concept studies within simplified dynamical systems, applying DL models to address specific challenges within the traditional DA framework. Notable examples include the Lorenz 63 and Lorenz 96 systems \cite{Lorenz1963,lorenz1996predictability}, as well as the 2-dimensional (2D) shallow water equations \cite{vreugdenhil2013numerical}. The scope of these applications encompasses the development of inverse observation operators \cite{frerix2021variational} for mapping observational data to physical states, the estimation of error covariance matrices \cite{cheng2022observation,melinc2023neural}, the joint training of surrogate models for the dynamic system along with DA problem solver \cite{brajard2020combining,arcucci2021deep,fablet2021learning}, and the derivation of tangent-linear and adjoint models \cite{hatfield2021building}.
Despite these achievements, a considerable gap exists between the simplified models used in these studies and actual weather systems in terms of the complexity and data dimensions.
This discrepancy underscores the urgent need for direct application of DL models to real-world weather systems. 
Recent studies have shown that DL models can match the forecasting prowess of NWP models with significantly lower computational costs \cite{bonavita20232022}, opening new avenues for DA development. However, the operational deployment of such DL models still rely on the analysis generated using traditional NWP models and DA systems. The optimization of this analysis data is for NWP model forecasts make it less ideal for DL-based weather forecasting models.
Therefore, there is a crtical need to design a DA system specifically tailored for these novel DL-based weather forecasting models. This systems should integrate real-world observations with the ultimate goal of establishing an operational framework that operates independently of traditional NWP and DA systems.


Given the pivotal role of satellite DA in improving the skill of medium-range weather forecasts \cite{geer2017growing,eyre2022assimilation}, this study introduces FuXi-DA, a pioneering, generalized DL-based DA framework. This framework is designed for the assimilation of satellite observations to produce the analysis that optimize the forecast performance of DL-based weather forecasting models.Figure 1 shows the comparison between FuXi-DA and the traditional DA process, taking the variational method as an example.
FuXi-DA advances beyond traditional DA systems by introducing several innovative features:
\begin{itemize}
\item To address the discrepancy in information content between background and observational data, FuXi-DA employs separate encoders for processing each type of data. This approach ensures the assimilation process fully leverages the extensive information contained in background data, while accurately integrating observational data. Notably, background data often surpasses observational data in spatial coverage and has more complete information on weather systems.
\item The pre-processing of observations involves a significant amount of laborious work, including data thinning, cloud detection, bias correction, and more \cite{eyre2022assimilation}. Such data pre-processing steps are not included in the FuXi-DA framework, or are implicitly integrated into the model architecture. Furthermore, it has been shown that FuXi-DA possesses the capability to implicitly learn from historical data to mitigate the impact of clouds on the assimilation of satellite data.
\item Traditional DA methods require the estimation of observation and background error covariance matrices, crucial for adjusting the influence of background and observational data within the DA system \cite{Lee2022}. FuXi-DA obviates this requirement through a innovative unified fusion neural network. This network effectively adeptly adjusts the weight of observations and background data without the need to estimate their error covariance matrices. This advancement addresses the challenges associated with the large size and impractical storage demands of the background error matrix, as well as the difficulty of accurate estimation due to the absence of a "true state."
\item Furthermore, FuXi-DA facilitates the joint training of DA with any DL-based weather forecasting model, specifically FuXi in this study, to not only refine analysis accuracy but also enhance medium-range forecast performance. This approach extends the forecast optimization window beyond the traditional assimilation window, a significant advancement over the traditional 4D-Var approach, which limits optimization to the observation availability window. Importantly, by leveraging supervised information from medium-range forecasts during training, FuXi-DA enables the simultaneous optimization of both assimilation and forecasting processes. This dual optimization significantly improves forecast accuracy.
\end{itemize}

In this study, we assess the efficacy of Fuxi-DA by integrating it with the Advanced Geosynchronous Radiation Imager (AGRI) aboard Fengyun-4B satellite. To illustrate the advantages and effectiveness of incorporating AGRI data, we developed a comparative correction model (refers to EXP\_CORR). This model replicates the architecture and training process of FuXi-DA (refers to EXP\_ASSI), with the only difference being its exclusion of AGRI data assimilation. Our results reveal that the assimilation of AGRI data consistently enhances forecast accuracy, demonstrating FuXi-DA's capability to leverage satellite data for forecast improvement. Further, multiple single-observation experiments confirm that FuXi-DA's consistency with established atmospheric physics.
It is important to note that while FuXi-DA was developed to refine the analysis for enhancing the forecast performance of FuXi, its framework is sufficiently flexible to be adapted for use with other DL-based weather forecasting models.

\begin{figure}[ht]
\centering
\includegraphics[width=1.0\textwidth]{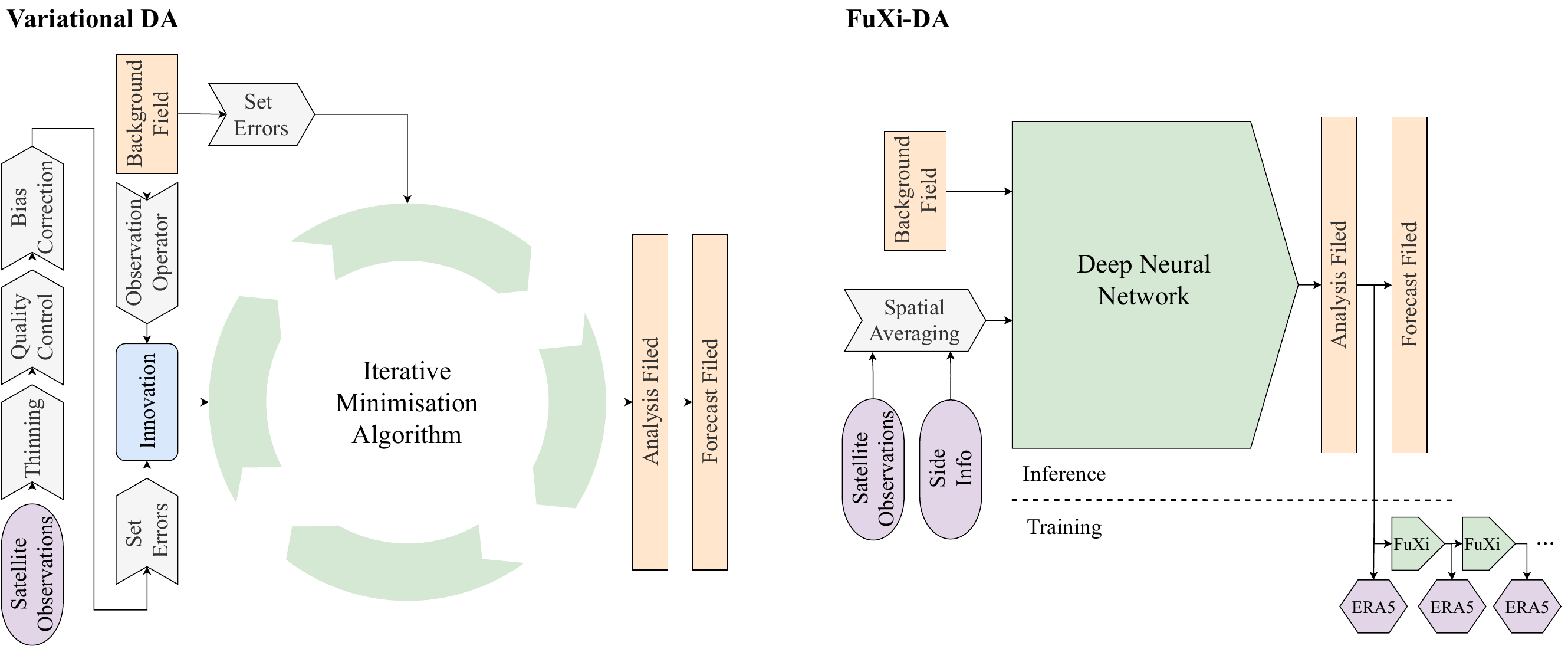}
\caption{The comparison between the Fuxi-DA and the variational data assimilation.}
\label{fig:framework}
\end{figure}

\section{Results}
Fengyun-4B/AGRI data, as the first selected satellite observation, have been successfully assimilated into FuXi-DA. Three experiments are designed to evaluate the impact of assimilating AGRI data in FuXi-DA, including performing a 6-hour forecast (called the background) using FuXi, correcting the 6-hour forecast fields using a correction model, and assimilating AGRI data into the 6-hour forecast fields using FuXi-DA (named EXP\_CTRL, EXP\_CORR and EXP\_ASSI, respectively). The EXP\_CORR experiment is designed for a fairer comparison, since the improvements in EXP\_ASSI experiment can be attributed to both the correction of the background data and the assimilation of AGRI data. The comparison between EXP\_CORR and EXP\_ASSI is the primary focus of the results.

After assimilating AGRI data in FuXi-DA, a significant reduction in analysis error is observed, particularly for relative humidity and geopotential. Furthermore, the impact of these observations is primarily evident in the middle and upper troposphere within the observed area. Specifically, the regionally-averaged latitude-weighted root mean square errors (RMSEs) of R300 and Z500 decrease by approximately 4.47\% and 2.02\%, respectively, within the coordinates of 52.5°E to 147.5°W and 80°S to 80°N. Additionally, the skill of global forecasts is improved after assimilating AGRI data. Within the forecast lead time of 7 days, there are statistically significant improvements, with the RMSE of Z500 decreasing by approximately 0.67\% at the forecast lead time of 1 days and 0.34\% at the forecast lead time of 7 days. We also demonstrated that the introduction of forecast fields loss improves the skill of long-term forecasting. Finally, FuXi-DA model exhibits excellent physical consistency and possesses the capability to automatically distinguish between observations under cloudy and clear conditions.


\begin{table}[htbp]
\begin{center}
\caption{The reduction rates of the regionally-average latitude-weighted RMSE for 65 upper-air atmospheric variables in the EXP\_ASSI experiment compared to the EXP\_CTRL experiment (left) and compared to the EXP\_CORR experiment (right).
}
\label{tab:region_rmse}
\begin{tabularx}{\textwidth}{c *{10}{C}}
\toprule
\multirow{2}{*}{\textbf{Pressure (hPa)}} & \multicolumn{5}{c}{\textbf{EXP\_ASSI compared to the EXP\_CTRL}} & \multicolumn{5}{c}{\textbf{EXP\_ASSI compared to the EXP\_CORR}} \\
\cmidrule(lr){2-6} \cmidrule(lr){7-11}
& \textbf{R} & \textbf{T} & \textbf{Z} & \textbf{U} & \textbf{V} & \textbf{R} & \textbf{T} & \textbf{Z} & \textbf{U} & \textbf{V} \\ \midrule
    1000 & -2.10 & -2.24 & -1.63 & -1.37 & -1.54 & -0.88 & -1.03 & -0.82 & 0.29 & 0.36  \\
    925 & -1.28 & -1.78 & -1.54 & -0.73 & -0.65 & -0.39 & -0.66 & -0.87 & 0.33 & 0.32  \\
    850 & -0.84 & -1.19 & -2.68 & -0.59 & -0.45 & -0.40 & -0.43 & -1.08 & 0.23 & 0.21  \\
    700 & -1.65 & -1.51 & -2.03 & -0.61 & -0.59 & -1.04 & -0.61 & -1.33 & -0.17 & -0.20  \\
    600 & -2.36 & -1.73 & -2.80 & -0.66 & -0.76 & -1.69 & -0.87 & -1.62 & -0.28 & -0.37  \\
    500 & -3.55 & -1.41 & -2.82 & -0.86 & -1.02 & -2.77 & -0.77 & -2.02 & -0.51 & -0.60  \\
    400 & -4.81 & -1.50 & -3.24 & -1.16 & -1.32 & -3.86 & -0.81 & -2.31 & -0.70 & -0.87  \\
    300 & -5.49 & -1.76 & -4.11 & -1.40 & -1.50 & -4.47 & -1.07 & -3.01 & -0.90 & -1.03  \\
    250 & -5.26 & -1.49 & -5.08 & -1.47 & -1.61 & -4.02 & -0.98 & -3.58 & -0.00 & -1.06  \\
    200 & -3.31 & -1.00 & -5.31 & -1.57 & -1.55 & -2.21 & -0.64 & -3.95 & -1.20 & -1.07  \\
    150 & -1.58 & -0.83 & -5.32 & -1.21 & -1.57 & -0.39 & -0.52 & -4.09 & -1.05 & -1.31  \\
    100 & -1.33 & -0.42 & -4.56 & -0.36 & -0.67 & -0.36 & -0.30 & -2.89 & -0.25 & -0.52  \\
    50 & -0.21 & -0.07 & -1.69 & -0.57 & -0.09 & -0.09 & -0.10 & -1.50 & -0.40 & -0.01  \\
\bottomrule
\end{tabularx}
\end{center}
\end{table}
\begin{figure}[htbp]
\centering
\includegraphics[width=1.0\textwidth]{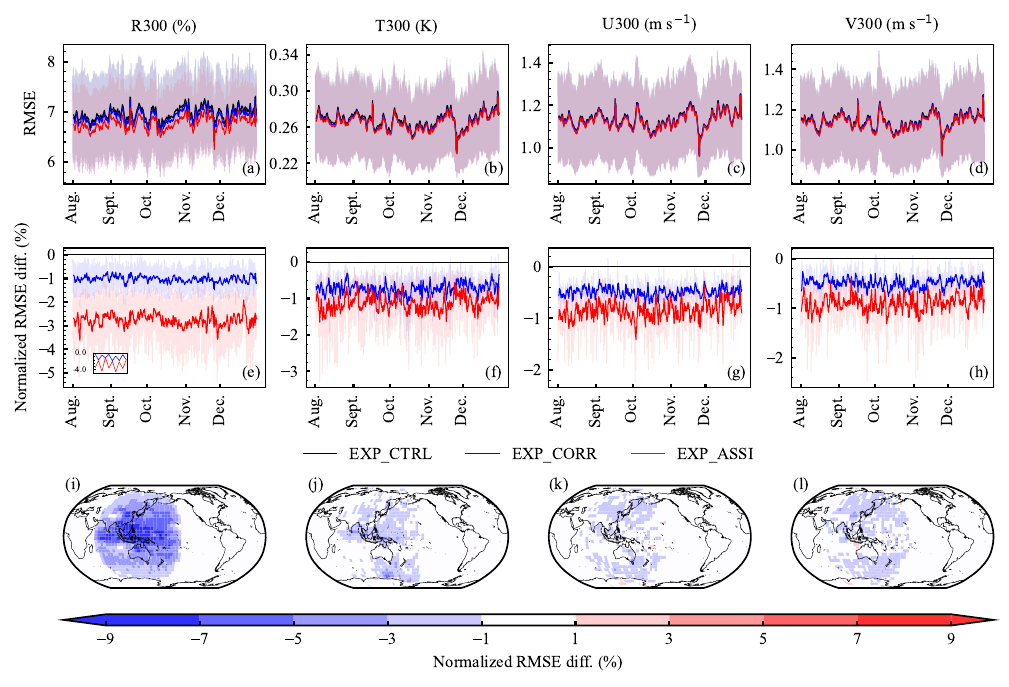}
\caption{Comparison of analysis after a single assimilation. The time series of the globally-averaged latitude-weighted RMSE (a, b, c, d) and the time series of the normalized RMSE difference (e, f, g, h) of EXP\_CTRL (black lines), EXP\_CORR (blue lines) and EXP\_ASSI (red lines), and the spatial distribution of normalized RMSE difference (i, j, k, l) for EXP\_ASSI relative to EXP\_CORR for R300, T300, U300 and V300. Sub-figure in (e) displays the original normalized RMSE difference at certain moments. The original data is represented by solid lines with high transparency, while solid lines with low transparency indicate the smoothed values.}
\label{fig:analysis}
\end{figure}
\subsection{Impact on the background}
This section compares the differences in background and analysis errors resulting from a single assimilation. The differences, within the observation region, between EXP\_CTRL and EXP\_ASSI are discussed first. The left side of Table \ref{tab:region_rmse} displays the reduction rates of the regionally-average latitude-weighted RMSE in the EXP\_ASSI experiment compared to the EXP\_CTRL experiment. As illustrated in the table, the assimilation of AGRI observations leads to a significant decrease in RMSEs across all variables, most notably in relative humidity and geopotential. To facilitate a more equitable comparison, the right side of Table \ref{tab:region_rmse} shows the comparison between the EXP\_ASSI and EXP\_CORR experiments. Significant improvements in relative humidity and geopotential are also observed in the middle and upper troposphere, with the RMSE of R300, R500, Z300, and Z500 decreasing by approximately 4.47\%, 2.77\%, 3.01\%, and 2.02\%, respectively. The notable improvement in humidity can be attributed to the three crucial water vapor channels of AGRI (channels 9-11). Additionally, there is a decrease in the RMSE of low-level temperatures, notably the 1000 hPa temperature, associated with the three window channels (channels 12-14).

Figure \ref{fig:analysis} shows the time series of RMSE and the normalized difference of globally averaged latitude-weighted RMSE for R300, T300, U300 and V300, as well as the horizontal distribution of the reduction rate of RMSE in the EXP\_ASSI experiment compared to EXP\_CORR experiment. The EXP\_CTRL, EXP\_CORR and EXP\_ASSI experiments are represented by black, blue and red colors, respectively. In Figure \ref{fig:analysis}a-\ref{fig:analysis}h, the original data is represented by solid lines with high transparency, while solid lines with low transparency indicate a moving average with a window size of 4. The spatial distribution of the normalized RMSE difference was calculated from regionally-averaged latitude-weighted RMSE over a range of 5° × 5°. As shown in Figure \ref{fig:analysis}a-\ref{fig:analysis}d, the RMSE values for EXP\_CTRL, EXP\_CORR and EXP\_ASSI experiments have consistently remained low. At almost all times, the EXP\_CORR and EXP\_ASSI experiments outperform the EXP\_CTRL experiment, exhibiting negative values in the normalized RMSE differences. Furthermore, throughout the long-term evaluation, all variable in the EXP\_ASSI experiment consistently perform better. Compared to the EXP\_CORR experiment, the reduction in RMSE within the observation area is more significant in the EXP\_ASSI experiment, which is consistent with our expectations. It is worth noting that the reduction in the RMSEs is more pronounced at 00 and 12 h (6-hour forecasts from 06 and 18), while the improvement is relatively small when assimilating at 06 and 18 h (6-hour forecasts from 00 and 12). At 00 and 12 h, there are typically more observations used in the 4D-Var component of ERA5 (for example, radiosonde soundings are often available at 00 and 12 h), resulting in higher analysis accuracy. Therefore, the background obtained from the 6-hour forecasts of FuXi at 00 and 12 h are more accurate. In this case, the assimilation model trusts the background more, leading to less impact from observations.

\begin{figure}[htbp]
\centering
\includegraphics[width=1.0\textwidth]{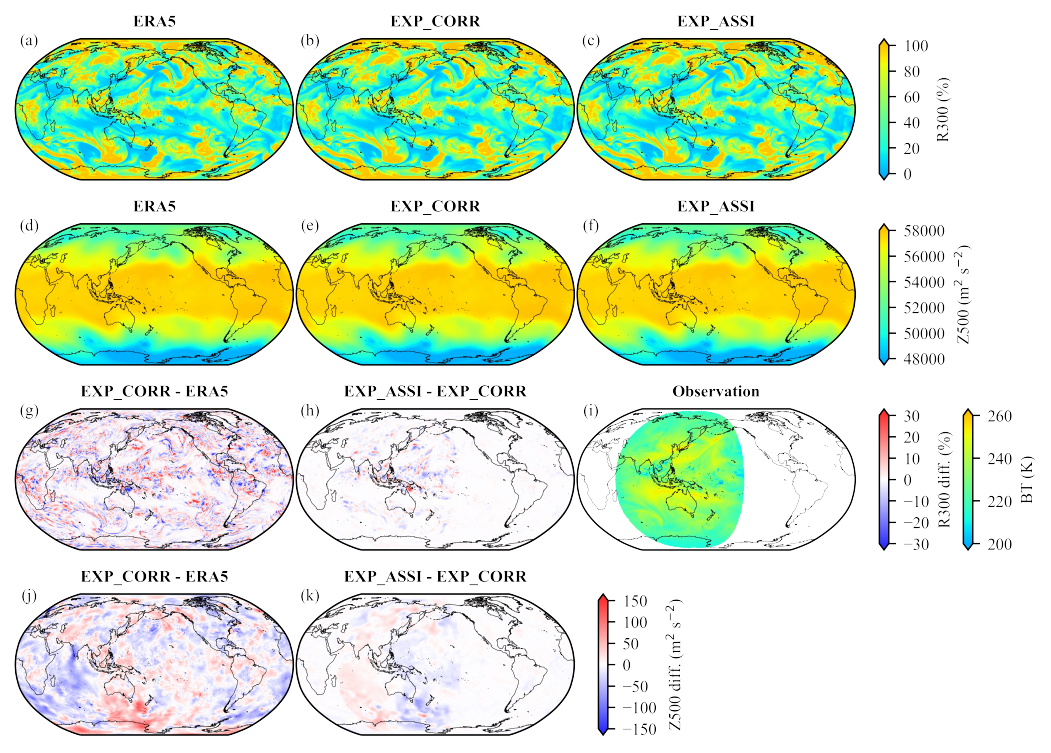}
\caption{Visualization of spatial distributions, error distributions and analysis increments of R300 and Z500, as well as visualization of observations. Spatial maps of R300 for ERA5 (a), EXP\_CORR (b), and EXP\_ASSI (c); spatial maps of Z500 for ERA5 (d), EXP\_CORR (e), and EXP\_ASSI (f); spatial maps of error distribution for R300 (g) and Z500 (j); spatial maps of analysis increments for R300 (h) and Z500 (k); and a spatial map of AGRI channel 9 brightness temperature (i).}
\label{fig:dxa}
\end{figure}

To further analyze the impact of assimilating AGRI observations, Figure \ref{fig:dxa} shows error distributions and analysis increments at 12:00 on October 15 2023. Figure \ref{fig:dxa}a-\ref{fig:dxa}f display the horizontal distributions of R300 and Z500 for ERA5, EXP\_CORR, and EXP\_ASSI, and there are no significant differences between them. Figure \ref{fig:dxa}g and \ref{fig:dxa}j show the errors between EXP\_CORR and ERA5, while Figure \ref{fig:dxa}h and \ref{fig:dxa}k display the errors between EXP\_ASSI and EXP\_CORR. As shown in Figure \ref{fig:dxa}h and \ref{fig:dxa}k, the analysis increments generated by assimilating AGRI data are concentrated primarily in the observation area (Figure \ref{fig:dxa}i). Furthermore, assimilating AGRI data leads to a reduction in analysis errors. For instance, in Figure \ref{fig:dxa}g, there is a significant negative deviation to the north of Australia. After assimilating AGRI data, a positive increment is generated (as seen in Figure \ref{fig:dxa}h). The improvement in geopotential is clearer. There are positive deviations to the east of Australia and negative deviations to the west of Australia (as seen in Figure \ref{fig:dxa}j). After assimilating AGRI, a negative increment was generated on the east side, and a positive increment on the west side (as seen in Figure \ref{fig:dxa}k).

\subsection{Impact on the forecast}
\begin{figure}[htbp]
\centering
\includegraphics[width=0.8\textwidth]{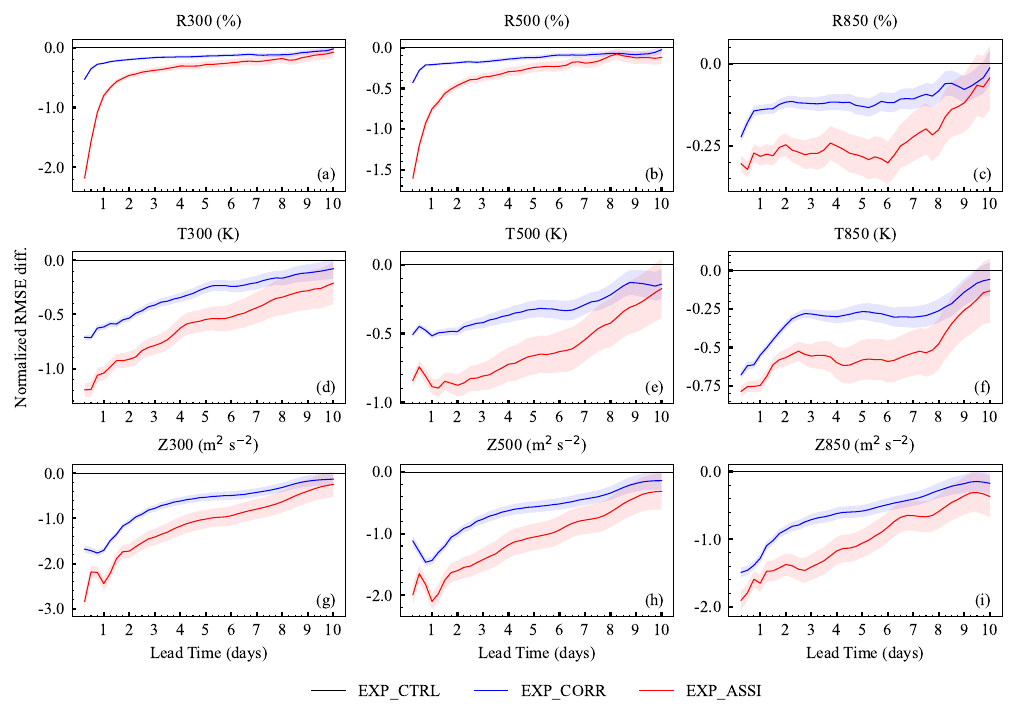}
\caption{Comparison of global 10-day forecasts after a single assimilation. The normalized RMSE difference of EXP\_CTRL (black lines), EXP\_CORR (blue lines) and EXP\_ASSI (red lines) for R300, R500 and R850 (a, b, c), T300, T500 and T850 (d, e, f), as well as Z300, Z500 and Z850 (g, h, i). The shaded areas represent 95\% confidence intervals.}
\label{fig:fcst}
\end{figure}

\begin{figure}[htbp]
\centering
\includegraphics[width=0.8\textwidth]{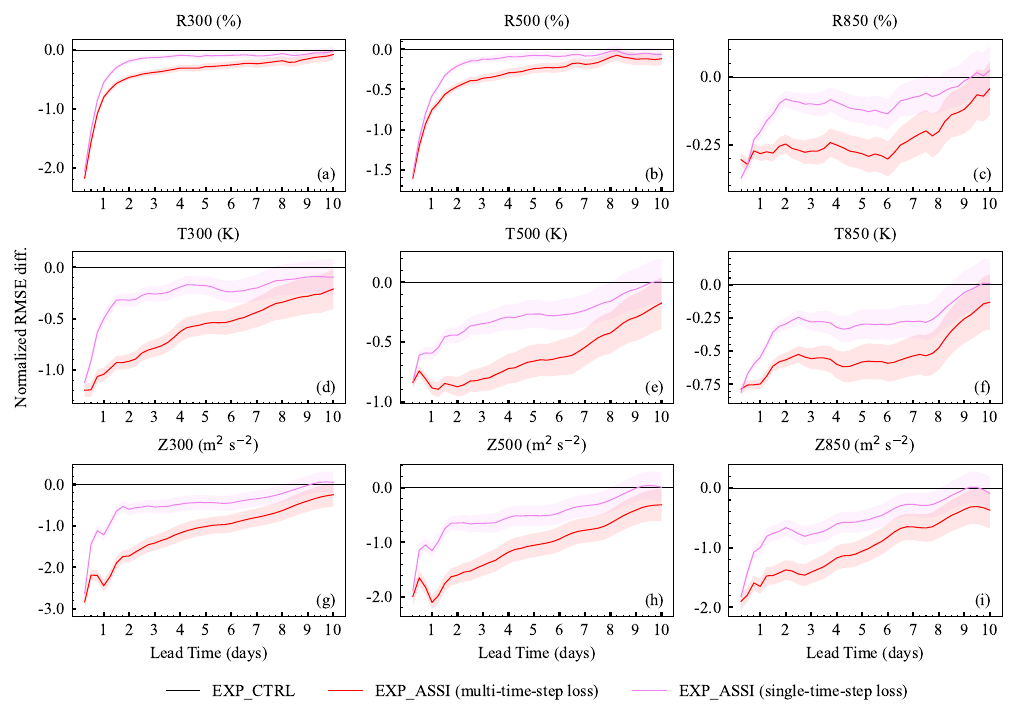}
\caption{Comparison of global 10-day forecasts after a single assimilation. The normalized RMSE difference of CTRL (black lines), EXP\_ASSI using a single-time-step loss (blue lines) and EXP\_ASSI using a multi-time-step loss (red lines) for R300, R500 and R850 (a, b, c), T300, T500 and T850 (d, e, f), as well as Z300, Z500 and Z850 (g, h, i). The shaded areas indicate a 95\% statistical confidence interval. The shaded areas represent 95\% confidence intervals.}
\label{fig:fcst_compare}
\end{figure}

Examining longer forecast ranges after a single assimilation, Figure \ref{fig:fcst} presents the normalized RMSE difference among EXP\_CTRL, EXP\_CORR and EXP\_ASSI for global 10-day forecasts across three variables at three pressure levels. The forecast skill is enhanced through correction and assimilation, and EXP\_ASSI performs better compared to EXP\_CORR. Similar to analysis, the improvement in relative humidity is most significant after assimilating AGRI observations. Observations' impact on forecasts is observed mainly in the middle and upper troposphere, exhibiting larger differences at 300 and 500 hPa, and smaller differences at 850 hPa between EXP\_CORR and EXP\_ASSI. With increasing forecast lead times, the improvements caused by assimilating AGRI data decrease gradually. Nevertheless, when compared to EXP\_CORR, EXP\_ASSI exhibits significant improvement in the first 7 days, as evidenced by the blue solid line extending beyond the red shading, and the reduction of Z500 errors ranges from 0.67\% at the forecast lead time of 1 day to 0.34\% at the forecast lead time of 7 days. It is worth mentioning that the accuracy of the forecast has been significantly enhanced through the joint optimization of forecasting and analysis, as illustrated in Figure\ref{fig:fcst_compare}. This improvement is likely attributable to the improved dynamic consistency of the initial fields resulting from the addition of multi-time-step forecast loss.

\begin{figure}[htbp]
\centering
\includegraphics[width=1.0\textwidth]{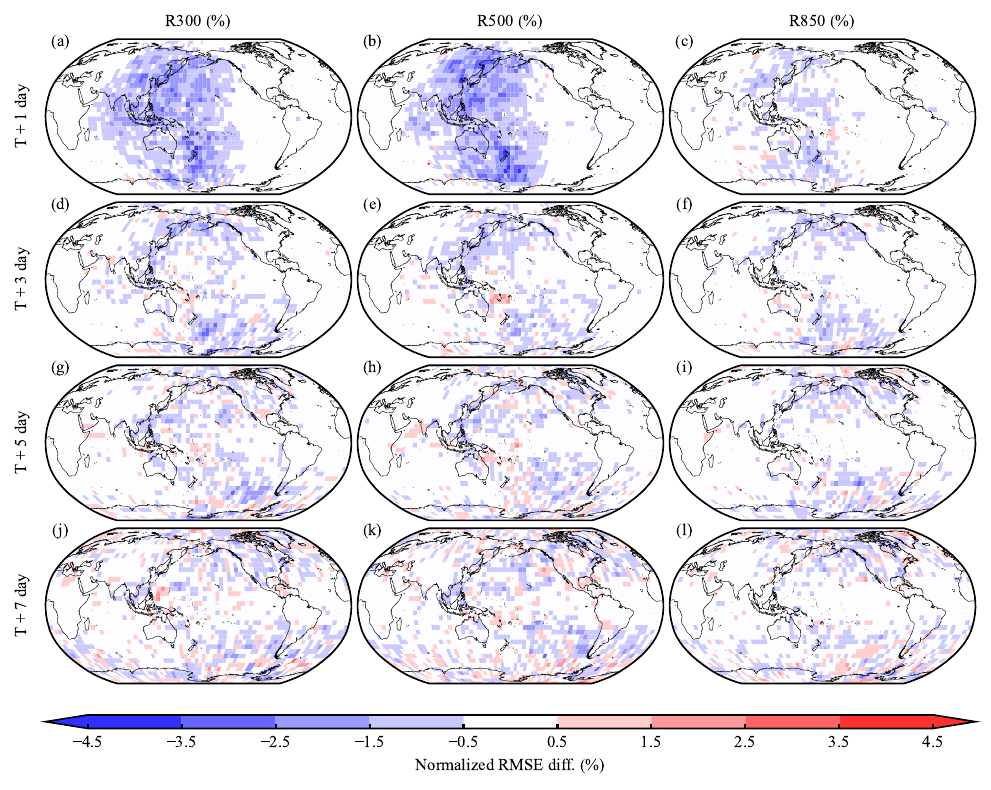}
\caption{Visualization of spatial distribution of normalized RMSE difference for humidity forecasts: EXP\_ASSI compared with EXP\_CORR. The first, second and third rows represent forecast lead times of 1 day, 3 days, 5 days and 7 days, respectively. The first, second and third columns represent pressures at 300, 500 and 850 hPa respectively. Statistics of latitude-weighted RMSE within a 5-degree by 5-degree range.}
\label{fig:fcst_pcolor}
\end{figure}

Figure \ref{fig:fcst_pcolor} depicts the spatial distribution of the impact from assimilating AGRI data on relative humidity forecasts. At the forecast lead time of 1 day, improvements in the humidity forecast are observed primarily within the observation areas. With increasing forecast lead times, the influence of assimilation expands rapidly. Driven by the mid-latitude westerlies, improvements in forecasting primarily propagate eastward. At the forecast lead time of seven days, the impact of assimilating AGRI data extends globally. Although there are negative effects in some regions for long-term forecasting, assimilating AGRI data demonstrates positive impacts in a greater number of regions.

\clearpage
\subsection{Interpretation}
Unlike previous works that utilized simulated observations for DL-based DA studies, we investigated whether physics-free DA networks are consistent with prior knowledge of atmospheric physics. Traditional DA methods frequently employ single-observation experiments to ascertain the validity of the impact generated by observations, which depends on the observation operator and the estimation of error in the background and observation. Here, we employed a similar approach. Given that the DA network lacks explicit observation operators and error statistics, and since using a single observation as input is not feasible in the trained FuXi-DA, we implemented some adjustments. The specific practice involved two inferences using the trained FuXi-DA. The first inference utilized the 6-hour forecast and original observations as inputs, while the second introduced a perturbation at a single point location to the original observations as model input. The difference between the analyses obtained from the two inferences is determined to characterize the analysis increment resulting from the perturbation. We begin with the traditional incremental analysis formula and provide a proof of this method in Appendix \ref{sec:single}.

\begin{table}[htbp]
\begin{center}
\caption{Observation perturbations test}
\label{tab:TB_perturb} 
\begin{tabularx}{\textwidth}{c *{3}{C}}
\toprule
\textbf{Channel} & \textbf{Condition} & \textbf{Brightness Temperature Perturbation} \\
\midrule
9 & clear   & +1 K \\ 
9 & clear   & +5 K \\
9 & clear   & -1 K \\
11 & clear  & +1 K \\
11 & cloudy   & +1 K \\ 
\bottomrule
\end{tabularx}
\end{center}
\end{table}

\begin{figure}[ht]
\centering
\includegraphics[width=0.8\textwidth]{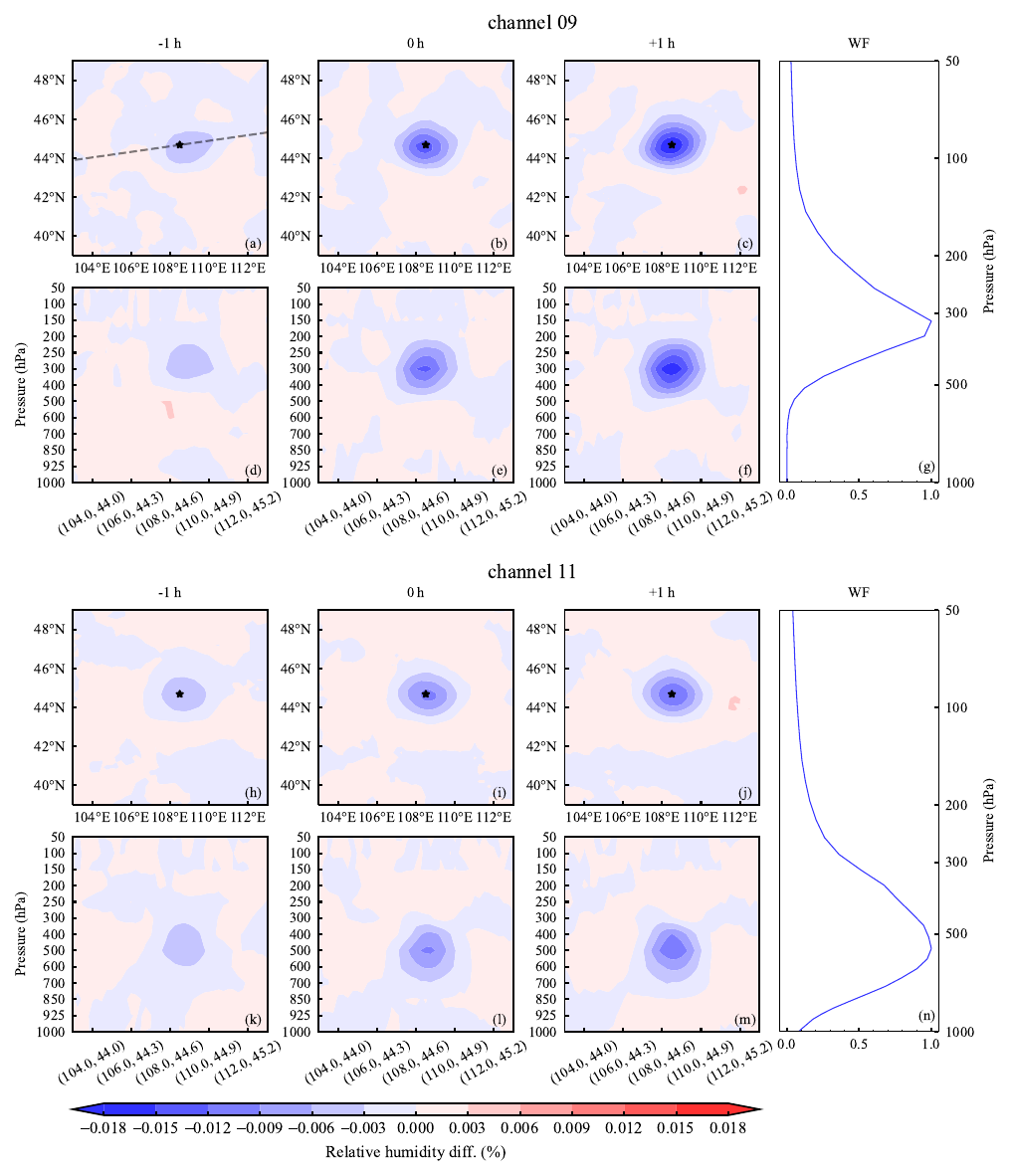}
\caption{The analysis increment of relative humidity resulting from introducing a perturbation of 1 K for the clear observation (pentagram) at the beginning (the first column), middle (the second column), and end (the third column) of the assimilation time window, respectively. Horizontal distribution of R300 increments resulting from introducing a perturbation of 1 K to channel 9 (a, b, c); vertical distribution of relative humidity increments resulting from introducing a perturbation of 1 K to channel 9 (d, e, f), obtained from a cross-section delineated by the dashed lines in (a); horizontal distribution of R500 increments resulting from introducing a perturbation of 1 K to channel 11 (h, i, j); vertical distribution of relative humidity increments resulting from introducing a perturbation of 1 K to channel 11 (k, l, m), obtained from a cross-section delineated by the dashed lines in (a); normalized weighting functions for channel 9 (g) and channel 11 (n). The atmospheric profile comes from the US standard atmosphere. The rapid radiative transfer model adopted is RTTOV version 13.2.}
\label{fig:single_clear}
\end{figure}

We randomly selected two observation points at 00:00 on October 26, 2023, one under clear conditions and the other under cloudy conditions (as shown in Figure \ref{single_points}.1). The distinction between clear and cloudy conditions was made using the AGRI Cloud Mask (CLM) product \cite{min2017developing}. Considering that relative humidity showed the most significant improvement after assimilating AGRI data, channels 9 (a high-level water vapor channel) and 11 (a low-level water vapor channel) were used for single-observation experiments. Table \ref{tab:TB_perturb} lists the five perturbation experiments conducted in this section. To further analyze the impact of introducing perturbations at different times within the assimilation window, perturbations were introduced at the beginning, middle, and end of the assimilation window, respectively. Figure \ref{fig:single_clear} shows the results of introducing a 1 K perturbation at the clear point. Figures \ref{fig:single_clear}a-\ref{fig:single_clear}c display the analysis increments of R300 resulting from perturbing channel 9. The decrease in relative humidity is consistent with the theory of radiative transfer in the atmosphere, indicating a negative correlation between brightness temperature and humidity. Specifically, a smaller amount of water vapor results in less absorption of surface-emitted radiation, thus the brightness temperature is warmer \cite{sieglaff2009inferring}. For a clear illustration, the humidity Jacobian functions for channels 9-11 are shown in Figure \ref{sec:WFJaco}.1 (a detailed description related to the Jacobian is provided in Appendix \ref{sec:WFJaco}). We also observed the performance of the increments under different perturbation values. By introducing a perturbation of -1 K, positive increments occurred, and the magnitude of these increments was consistent with those caused by adding a perturbation of 1 K (as shown in Figure \ref{single_points}.2). However, introducing a perturbation of 5 K resulted in the increments of relative humidity being approximately five times larger than those from a perturbation of 1 K. These phenomena align with the results of traditional DA methods. Figures \ref{fig:single_clear}h-\ref{fig:single_clear}m show the analysis increments of R500 resulting from introducing a perturbation to channel 11, displaying results similar to those of channel 9. It is important to note that the characteristic of the analysis increments is the dispersion from the observation point towards the surroundings, which is caused by the spatial correlations of the background.

\begin{figure}[ht]
\centering
\includegraphics[width=0.8\textwidth]{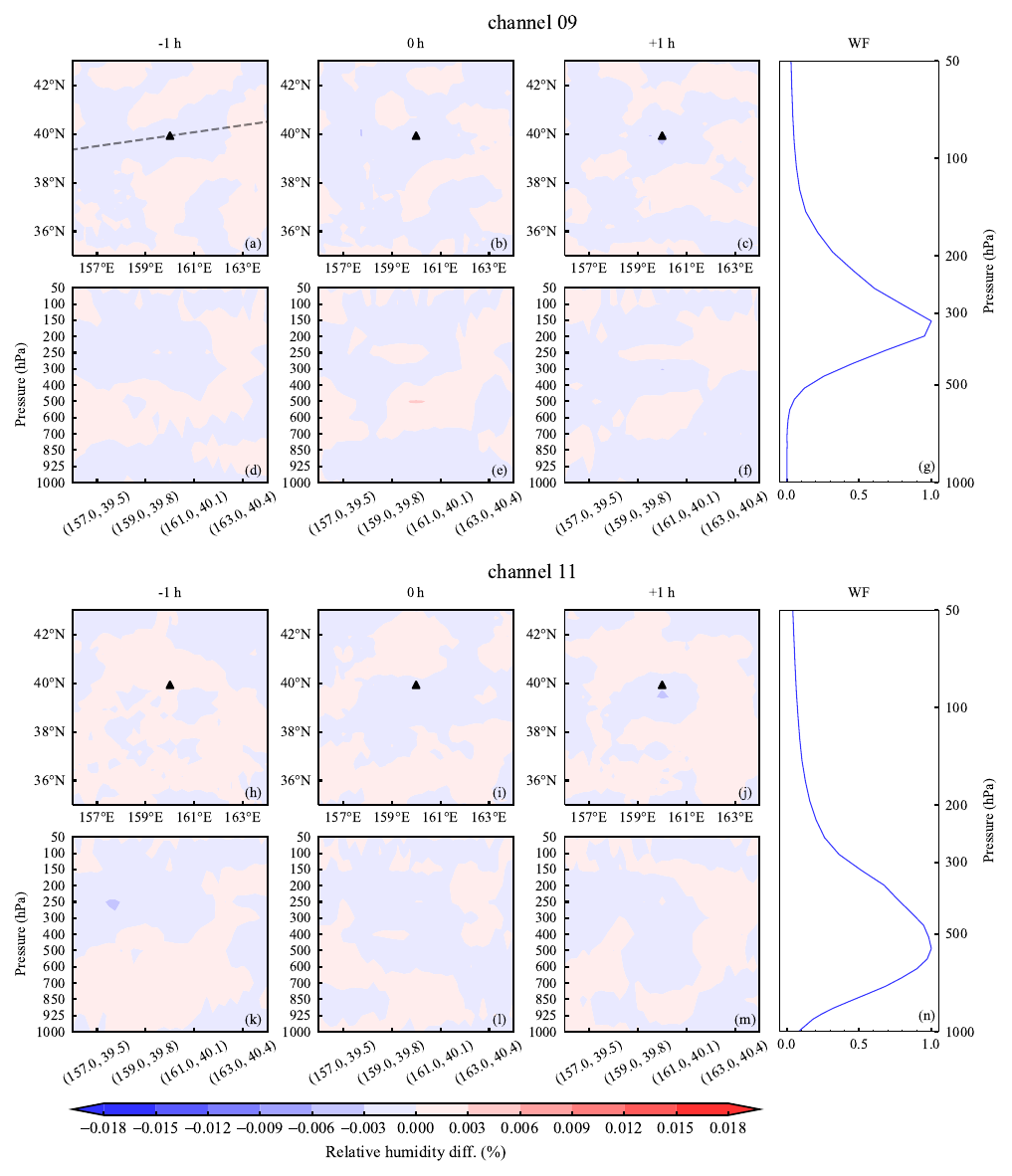}
\caption{Similar to Figure \ref{fig:single_clear} but for cloudy observation (pentagram).}
\label{fig:single_cloudy}
\end{figure}

To further demonstrate physical consistency, Figure \ref{fig:single_clear} also shows the vertical distribution of the relative humidity increment. The analysis increment of channel 9 reveals the vertical correlation of the background, with the peak increment occurring around 300 hPa (as shown in Figures \ref{fig:single_clear}d-\ref{fig:single_clear}f). Figure \ref{fig:single_clear}g shows the normalized weighting functions for channel 9 (a detailed description of the weighting function is provided in Appendix \ref{sec:WFJaco}). It can be noted that the peak of the analysis increment aligns with the peak of the weighting function, both located around 300 hPa. Additionally, the findings for channel 11 are similar to those for channel 9, except that the peak increment for channel 11 occurs at a lower altitude, around 500 hPa (as shown in Figures \ref{fig:single_clear}k-\ref{fig:single_clear}m). These results demonstrate that the impact of observations on the assimilation results is correct, further demonstrating the stability of FuXi-DA. Notably, the analysis increments are very small after introducing a 1 K perturbation. One reason is that the observations were not sparsified, so adjacent observations can be correlated. When multiple observations jointly impact the same location in the background, the expected analysis increment will be larger. Another possible reason is the high accuracy of the 6-hour forecast field of FuXi, leading the assimilation model to trust the background more. Additionally, all perturbation experiments show an interesting phenomenon: the closer to the end of the assimilation window, the greater the impact of observations. The same phenomenon appears in the traditional 4D-Var assimilation system \cite{mcnally2019sensitivity}. In fact, the assimilation method of FuXi-DA is more similar to the the three-dimensional variational (3D-Var) method. However, by incorporating the temporal information of the observational data, FuXi-DA has demonstrated phenomena consistent with the 4D-Var method, which is encouraging.
Traditional satellite DA methods require complex modeling for assimilating observations under cloudy conditions \cite{bauer2010direct, eyre2022assimilation, li2022satellite}. Additionally, many cloud-affected observations from instruments that have not achieved all-sky assimilation are discarded. In this study, the FuXi-DA model assimilates observations under both cloudy and clear conditions. To demonstrate the adaptive adjustment ability of the DA network under different conditions, a cloudy single-observation experiment was conducted. Figure \ref{fig:single_cloudy} shows the analysis increment of relative humidity resulting from introducing a 1 K perturbation to channels 9 and 11 at the cloudy observation point (as shown in Figure \ref{single_points}.1). There is no significant analysis increment around the cloudy observation point, which indicates that FuXi-DA possesses autonomous cloud detection capabilities. The limited impact of cloudy observations on the background is due to clouds having a strong ability to absorb infrared radiation, making it difficult to obtain information below the clouds. Additionally, the lack of hydrometeors (such as liquid, ice, snow, hail, graupel) and other particles (including mixed phase) in the background makes it difficult for the network to establish a connection between the observation and the background. The varied performance of FuXi-DA under clear and cloudy conditions demonstrates the potential of the DL model to achieve all-sky assimilation.  
\section{Discussion}
Both numerical weather prediction (NWP) systems and the recently developed deep learning-based weather forecasting models rely on accurate initial conditions. Modern NWP systems utilize data assimilation (DA) to determine the current state of the atmosphere. This process is very complex and requires significant computational resources. Currently, satellite observations have become the dominant source of observational information in DA systems, but assimilating these data is more complex and consumes more computational resources. Therefore, we have developed a scalable satellite DA framework based on deep learning (DL) models, named FuXi-DA. FuXi-DA employs different encoders for various types of data, resolving the inconsistency between observational and background variables by facilitating interactions among diverse data in latent space. Starting from the general form of traditional variational methods \cite{lorenc1986analysis}, FuXi-DA automatically learns the weights for observations and backgrounds in a unified fusion neural network to facilitate the assimilation process. Using the FuXi-DA assimilation framework, we have successfully implemented the assimilation of Fengyun-4B/AGRI data.

Results indicate a significant improvement in analysis and forecast accuracy after assimilating AGRI data. The successful assimilation of AGRI data in FuXi-DA represent the first step towards a fundamental shift from traditional DA methods to DL-based DA models. FuXi-DA's strong performance in fusing background and observation information, its design simplicity and scalability, the rapidity of computation, and the unique approach to handling observation data create a possibility for building an end-to-end DL-based weather forecasting system. Additionally, FuXi-DA demonstrates consistency with prior knowledge of atmospheric physics, enhancing confidence in the application of multi-source observations in DL-based DA models. 

Compared to traditional DA systems, FuXi-DA not only significantly simplifies the data pre-processing steps but also substantially reduces the computational resources required. In traditional DA systems, specific modules such as data thinning, quality control, bias correction, error estimation, and observation operators need to be customized for different observations, consuming significant human and computational resources. As mentioned above, FuXi-DA avoids the manual construction of observation operators and error estimation by encoding each type of data separately and employing a unified fusion module. Furthermore, FuXi-DA does not require data thinning and cloud detection operations, thus maximizing the utilization of observations. Although it has not demonstrated the capability to assimilate observational data under cloudy conditions, FuXi-DA effectively distinguishes observations between clear and cloudy conditions. Introducing hydrometeors and other particles into the forecast model could further enhance the capability of all-sky assimilation. Additionally, we incorporate satellite zenith angle and spatio-temporal information into the inputs of FuXi-DA, and adjust observations within the fusion module to achieve bias correction functionality. In summary, FuXi-DA reduces pre-processing steps through its strong learning capability and achieves the necessary processes through its unique modeling approach. Most importantly, FuXi-DA greatly saves computational resources. Compared to the extensive computational resources consumed by traditional DA systems \cite{bauer2015quiet}, FuXi-DA completes one assimilation on a single A100 GPU in less than 10 seconds.

Looking to the future, our goal is to build an end-to-end DL-based weather forecasting system. To achieve this objective, we are endeavoring to incorporate more satellite observation data into FuXi-DA, including advanced infrared sounders, microwave sounders, and imagers, among others \cite{eyre2022assimilation}. Furthermore, we will continue to explore the application of more unevenly distributed and sparse observation data in DL-based assimilation models, such as radiosonde soundings and aircraft meteorological data reports \cite{moninger2003automated,ingleby2016progress}. Additionally, considering that only one year of data were used for model training, which is significantly less compared to the amount of data typically used in training DL forecast models, combining real observations with simulated observations for joint training is expected to yield better results in future work.

\section{Method}
\label{sec:method}

\subsection{Data}
\label{sec:data}

ERA5 is the fifth generation of the ECMWF reanalysis dataset, generated by assimilating high-quality and abundant global observations using ECMWF’s IFS model. The ERA5 data is widely regarded as the most comprehensive and accurate reanalysis archive. Therefore, we use the ERA5 reanalysis dataset as the ground truth for model training. Moreover, ERA5 data is also used as the input for generating the background in the FuXi model. In this work, we use 5 upper-air atmospheric variables at 13 pressure levels (50, 100, 150, 200, 250, 300, 400, 500, 600, 700, 850, 925, and 1000 hPa), and 5 surface variables. The 5 upper-air atmospheric variables are geopotential (Z), temperature (T), u component of wind (U), v component of wind (V), and relative humidity (R). Additionally, 5 surface variables are T2M, 10-meter u wind component (U10), 10-meter v wind component (V10), mean sea level pressure (MSL), and total precipitation (TP). In total, 70 variables are used in this study, with a spatial resolution of 0.25° and a temporal resolution of 6 hours.

Fengyun-4B/AGRI has a total of 15 channels, including 3 visible (VIS) channels, 3 near-infrared (NIR) channels, 2 mid-wave infrared (IR) channels, 3 water vapor channels, and 4 long-wave IR channels. The footprint at nadir for the VIS channels is 1 km, for the NIR channels and one mid-wave IR channel it is 2 km, and for the remaining channels, it is 4 km \cite{AGRI_observation}. In our research, observations from channels 8-15 are used for assimilation, with an assimilation time window of plus or minus one hour. Since the footprint of these channels is 4 km (2748 × 2748 latitude-longitude scattered points), and the background fields have a spatial resolution of 0.25° (721 × 1440 latitude-longitude grid points), achieving correspondence of the respective positions within the network for these two types of data poses a challenge. We processed the satellite observations into "super-observations" that matched the resolution of the background fields. This approach makes the observations more representative of the spatial scales in the model and effectively reduces the uncertainty of the observations \cite{geer2010enhanced, bell2008assimilation}. The specific process includes: (1) Cropping a regular grid area that includes the observation area based on the background coordinates. For AGRI data, the corresponding background size is approximately 640 × 640. (2) Averaging the satellite observations within ±0.125° around each grid point of the cropped area to represent the satellite observations at that grid point. (3) Applying mask processing to positions within the cropped area that do not have matched observations. After processing, the satellite observations were processed into 640 × 640 latitude-longitude grid point data with a spatial resolution of 0.25°.

The data for FuXi-DA is provided in the form of input-output pairs, where the inputs include background fields and satellite observations, and the output is ERA5 data. The available data spans a period from June 2022 to December 2023. The data is divided into training, validation, and testing sets. The training includes 1460 samples spanning from June 2022 to May 2023. The validation set contains 244 samples, corresponding to the period from June 2023 to July 2023, while out-of-sample testing is performed using 612 samples from August 2023 to December 2023.

\subsection{Model and Architecture}
\label{sec:architecture}

In the three-dimensional variational (3D-Var), the general form for the analysis can be expressed as:

\begin{equation}
x^a=x^b+BH^T(HBH^T+R)^{-1}[y^o-H(x^b)]
\label{eq1}
\end{equation}

where $x^a$ is the state vector of analysis field, $x^b$ is the state vector of background filed, $y^o$ is the observation vector, $H$ is the observation operator, $B$ is the covariance matrix of background error and $R$ is the covariance matrix of observation errors, see e.g. \cite{lorenc1986analysis} for this standard result. The construction of FuXi-DA was inspired by this form. We depict DA learning as an incremental learning process, where FuXi-DA learns analysis increments from a fusion module and applies them to the background. 

\begin{figure}[ht]
\centering
\includegraphics[width=0.85\textwidth]{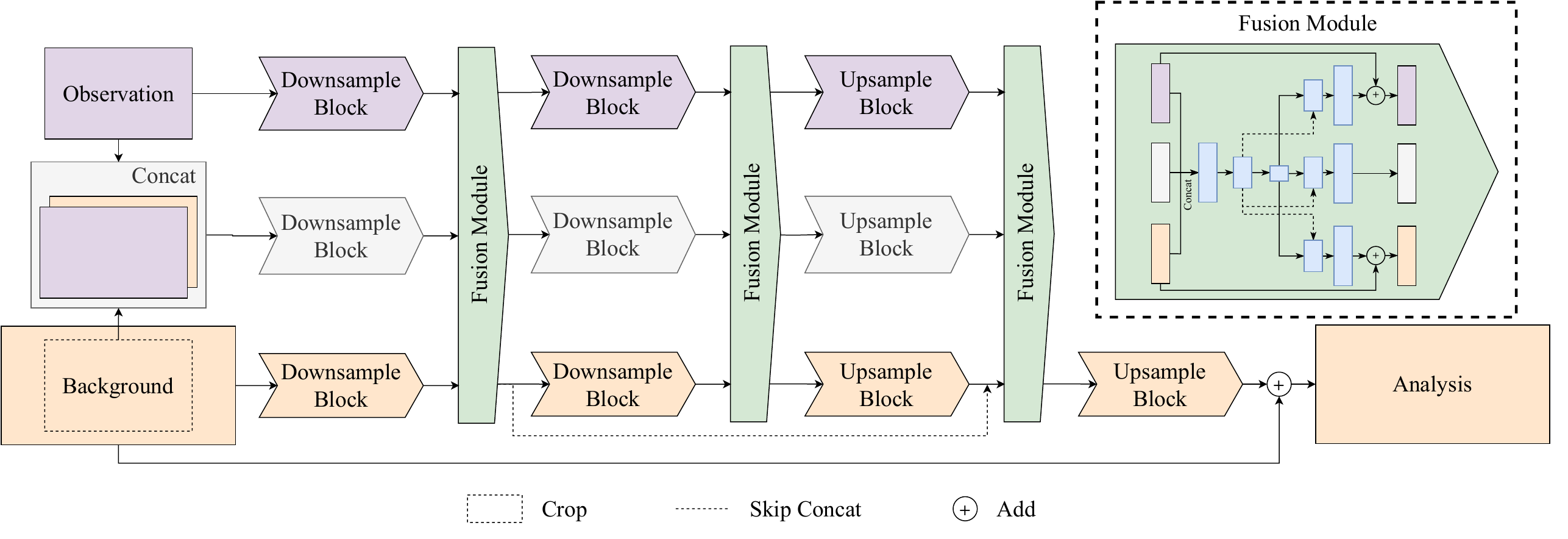}
\caption{Overall architecture of FuXi-DA model.}
\label{fig:architecture}
\end{figure}

Due to the inconsistency between the background field variables and observational variables, we encode each input separately and implement interactions among various data within a unified fusion model. Compared to the traditional DA system, which interpolates background fields to the position and time of each observation and then processes them through an observation operator for comparison with observations \cite{eyre2022assimilation}, FuXi-DA unifies and simplifies this process by converting both observations and background into the latent space. This approach significantly saves human and computational resources and avoids errors introduced by the observation\cite{watts2004identification, auligne2007adaptive}. As shown in Figure \ref{fig:architecture}, the model architecture of the FuXi-DA consists of three branches. The three branches are the background information flow (yellow boxes), the mixed information flow (gray boxes), and the observation information flow (violet boxes), respectively. All branches adopt the same U-net architecture \cite{ronneberger2015u}. The background flow has two down-sample and two up-sample stages, while the others has two down-sample stages and one up-sample stages. The channels for each U-net stage are 256, 512 and 256 respectively.

Firstly, the input data for FuXi-DA includes the background, observations, and additional observation information. The background has dimensions of 70 × 721 × 1440, where 70, 721, and 1440 represent the total number of input variables, latitude and longitude grid points, respectively. In FuXi-DA, the background input is first reshaped into 70 × 720 × 1440 through bilinear interpolation. The processed observations have dimensions of 8 × 8 × 640 × 640, where 8, 8, 640, and 640 represent the total number of time frames, channels, latitude and longitude grid points, respectively. Considering that satellite observation biases usually vary with scan angle and geographic location \cite{harris2001satellite, auligne2007adaptive, yin2020evaluation}, we encode the longitude, latitude, satellite zenith angle, and temporal information of the satellite observations. The longitude, latitude, and satellite zenith angle are taken as the cosine, sine, and cosine, respectively. For time encoding, the observation time is first transformed into the day of the year and the minute of the day. Then, we calculate both the cosine and sine for the day of the year and the minute of the day. By encoding longitude, latitude, satellite zenith angle, and time and adding them as additional observation channels in the model, the size of observations becomes 8 × 15 × 640 × 640. The observation input is first reshaped into 120 × 640 × 640. Then, the background input is cropped to correspond to the positions of the observations and concatenated with the reshaped observations to obtain the input of the mixed information flow, with a size of 190 × 640 × 640. Following the model architecture illustrated in Figure \ref{fig:architecture}, the output is initially reshaped to 70 × 720 × 1440, then restored to the original input shape of 70 × 721 × 1440 by bilinear interpolation.

The fusion module receives information from the three branches and facilitates information interaction through the U-net module with two down-sample and two up-sample stages. Each fusion module has three outputs. The first output is added to the background information flow. Similar to how traditional DA systems adjust the weights of the background and observations through background errors and observation errors, FuXi-DA automatically learns the weights of the background and observations within this module and outputs information that implicitly contains analysis increments. The second output is propagated as mixed information to the subsequent network. The third output is added to the observation information flow. This operation draws on the traditional bias correction process, where predictions computed from the background field are typically used for bias correction \cite{eyre1992bias,harris2001satellite,auligne2007adaptive}. 

In Fuxi-DA, each down-sample stage consists of a 2 × 2 2-dimensional (2D) convolution layer with a stride of 2, a layer normalization layer \cite{ba2016layer}, a sigmoid-weighted linear unit (SiLU) activation \cite{elfwing2018sigmoid} and a 3 × 3 2D convolution layer with a stride of 1. Each up-sample stages consists of a 3 × 3 2D convolution layer with a stride of 1, a layer normalization layer, a sigmoid-weighted linear unit (SiLU) activation, a 3 × 3 2D convolution layer with a stride of 1 and a pixel-shuffle layer \cite{shi2016real} with an upscaling factor of 2. Furthermore, a skip connection is included that concatenates the outputs from the down-sample stage with the output of up-sample stage before being fed into the next up-sample stage.

\subsection{Model training}
\label{sec:training}
The FuXi-DA model is built using the PyTorch framework \cite{paszke2017automatic}. The model was trained on 4 Nvidia A100 GPUs for a total of 6000 iterations, with a batch size of 1 per GPU. The entire model training process lasted about 8 hours. The AdamW \cite{loshchilov2017decoupled} optimizer is used with parameters $\beta_1$ = 0.9, $\beta_2$ = 0.999 and a weight decay coefficient of $1e^{-5}$. The learning rate was set using a warmup and a cosine annealing schedule \cite{loshchilov2016sgdr}. First, the learning rate increases during the training process according to the formula:
\begin{equation}
lrate=[(step\_num-1)/warmup\_steps]\times(stop\_lrate-start\_lrate)+start\_lr
\label{eq2}
\end{equation}
This corresponds to increasing the learning rate linearly for the first warmup\_steps training steps. We used $warmup\_steps=500$, $start\_lrate=1e^{-8}$, and $stop\_lrate=2e^{-3}$. Second, the learning rate decreases through a cosine annealing schedule with an initial learning rate of $2e^{-3}$ and maximum number of iterations of 6000. 
We use the latitude-weighted $L1$ loss to minimize the errors between Fuxi-DA's outputs and the ERA5, which is defined as follows: 

\begin{equation}
L1\_Loss=
\frac{1}{C\times{H}\times{W}}
\displaystyle\sum_{c=1}^C
\displaystyle\sum_{i=1}^H
\displaystyle\sum_{j=1}^W
\alpha_i|\widehat{X}_{c,i,j}-X_{c,i,j}|
\label{eq3}
\end{equation}

where $C$, $H$ and $W$ are the number of channels and the number of latitude and longitude grid points, respectively. $\widehat{X}$ is the ground truth. $\alpha_i= H\times\left.{cos\Phi_i}\middle/ \right.{\displaystyle\sum_{i=1}^Hcos\Phi_i}$ is the weight at latitude $\Phi_i$. 
In the actual training process, we employed a multi-time-step loss. The output of FuXi-DA is used as the input for FuXi for long-range forecasting, and the errors between the forecast results and ERA5 need to be minimized. The final loss function used in Fuxi-DA is defined as follows: 

\begin{equation}
Total\_Loss= L1\_Loss^0 + 
\frac{1}{T}
\displaystyle\sum_{t=1}^T L1\_Loss^t
\label{eq4}
\end{equation}

$T$ represents the number of time steps, where at $t=0$, $X$ denotes the current analysis field, and at $t>0$, $X$ represents the forecast field. In this study, we set $T=10$, indicating the inclusion of the analysis field and 10 forecast fields spanning 6-60 hours.

\subsection{Experience setup}
\label{sec:expset}

Three experiments were designed to assess the impact of assimilating AGRI data on analysis and forecasts. The control experiment, referred to as “EXP\_CTRL,” pertains to the 6-hour forecast results from FuXi. Given that FuXi was trained with data from 1979 to 2015, biases are present in the 6-hour forecasts during the study period of this work. Consequently, the output of FuXi-DA is influenced by both assimilation and correction. To render the results more rigorous, we trained a correction model (the corresponding experiment is named "EXP\_CORR"). The training dataset for the correction model is consistent with that used to train FuXi-DA, except that satellite observations were excluded. Additionally, the model structure adopts a similar architecture to FuXi-DA, but with the fusion module removed, while the training approach is completely consistent with FuXi-DA. The assimilation experiment is named EXP\_ASSI.

\subsection{Evaluation method}
\label{sec:Eval}

The Latitude- weighted root mean square error (RMSE) are used to evaluate analysis and forecast performance, which is calculated as follows:
\begin{equation}
RMSE(c)=\sqrt{
\frac{1}{H\times{W}}
\displaystyle\sum_{i=1}^H
\displaystyle\sum_{j=1}^W
\alpha_i(\widehat{X}_{c,i,j}-X_{c,i,j}})^2
\label{eq5}
\end{equation}
Additionally, The normalized RMSE difference between experiment A and experiment B calculated as $(RMSE_A-RMSE_B)/RMSE_B$. Similarly, the regionally-average latitude-weighted RMSE is calculated as follows:
\begin{equation}
RMSE(c)=\sqrt{
\frac{1}{(H_{max}-H_{min})\times({W_{max}-W_{min}})}
\displaystyle\sum_{i=H_{min}}^{H_{max}}
\displaystyle\sum_{j=W_{min}}^{W_{max}}
\widehat{\alpha}_i(\widehat{X}_{c,i,j}-X_{c,i,j}})^2
\label{eq6}
\end{equation}
where $H_{min}$, $H_{max}$, $W_{min}$ and $W_{max}$ represent the range of grid point indices corresponding to the selected region. $\widehat{\alpha}_i= (H_{max}-H_{min})\times\left.{cos\Phi_i}\middle/ \right.{\displaystyle\sum_{i=H_{min}}^{H_{max}}cos\Phi_i}$ is the weight at latitude $\Phi_i$. 


\bibliographystyle{unsrt}  
\bibliography{references}

\newpage

\appendix

\renewcommand{\theequation}{\thesection.\arabic{equation}} 
\setcounter{equation}{0}

\renewcommand\thefigure{\thesection.\arabic{figure}} 
\setcounter{figure}{0} 

\section*{Appendix}

\section{Single-observation experiments}
\label{sec:single}

According to Equation \ref{eq1}, the analysis can be simplified as:

\begin{equation}
x^a=x^b+K[y^o-H(x^b)]
\label{eq7}
\end{equation}

Here, $K=BH^T(HBH^T+R)^{-1}$ is Kalman gain matrix. If a perturbation $\Delta y$ is added to $y^o$, the new analysis $x^{a'}$ can be expressed as:

\begin{equation}
x^{a'}=x^b+K[y^o+\Delta y-H(x^b)]
\label{eq8}
\end{equation}

From equations \ref{eq7} and \ref{eq8}, the analysis increment $\Delta x^{a'}$ resulting from the perturbation can be expressed as:

\begin{equation}
\Delta x^{a'}=K[\Delta y]
\label{eq9}
\end{equation}

\begin{figure}[ht]
\centering
\includegraphics[width=1.0\textwidth]{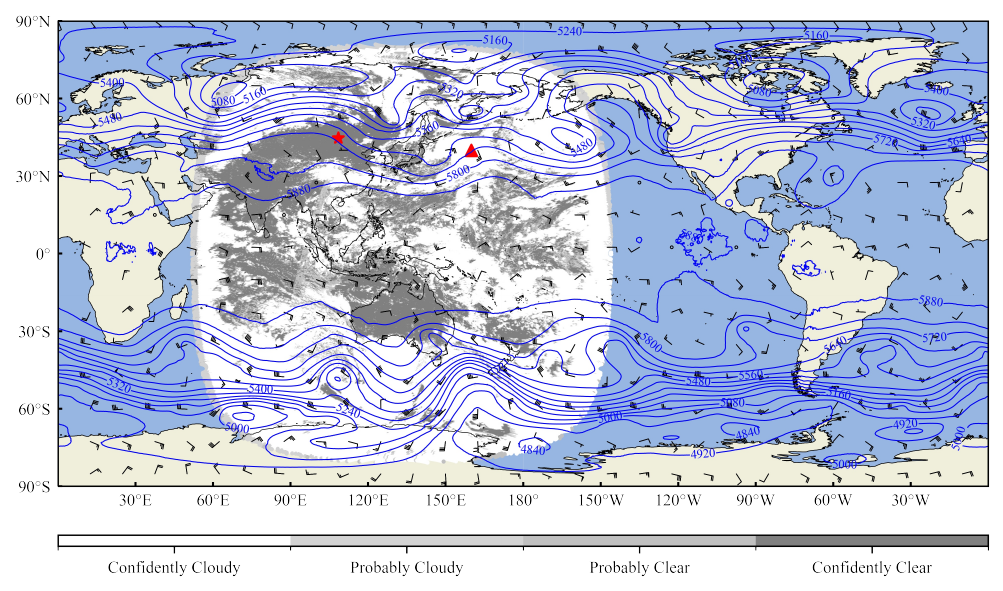}
\label{single_points}
\caption{Visualization of spatial distributions of AGRI CLM product and the selected observation locations (pentagram and triangle) at 00:00 on October 26, 2023. Blue solid lines represent the 500 hPa geopotential height contours of the background at intervals of 80, and black wind shafts indicate the 850 hPa wind vectors of the background (a half barb represents a wind speed increment of 2 m/s, a full barb represents a wind speed increment of 4 m/s and a flag represents a wind speed increment of 20 m/s).}
\end{figure}

\begin{figure}[ht]
\centering
\includegraphics[width=1.0\textwidth]{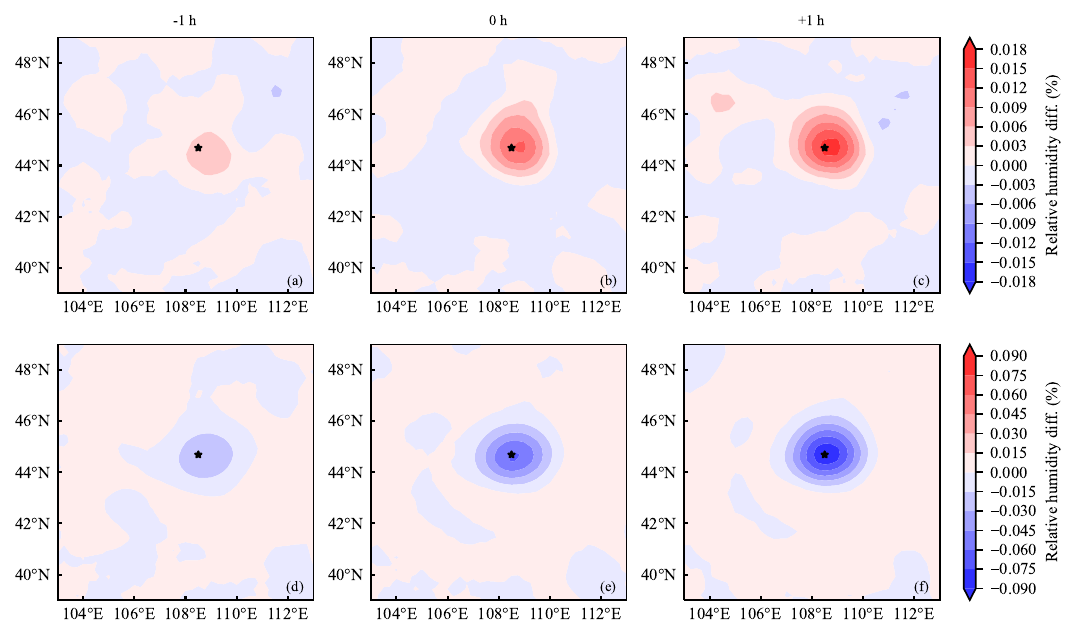}
\caption{The analysis increment of relative humidity resulting from introducing a perturbation of 1 K (a, b, c) and 5 K (d, e, f) to channel 9 for the clear observation (pentagram) at the beginning (the first column), middle (the second column), and end (the third column) of the assimilation window, respectively.}
\label{fig:single_other}
\end{figure}

\renewcommand{\theequation}{\thesection.\arabic{equation}} 
\setcounter{equation}{0}

\renewcommand\thefigure{\thesection.\arabic{figure}} 
\setcounter{figure}{0} 

\section{Weight Function and Jacobian}
\label{sec:WFJaco}

The radiative transfer model (RTM) RTTOV version 13.2 is used as the observation operator to simulate the clear-sky radiances and the atmospheric profile comes from the US standard atmosphere. RTTOV is developed by the Satellite Application Facility on Numerical Weather Prediction (NWP SAF) of EUMET- SAT. The weighting function (WF) of a channel with central wavenumber $v$ in a satellite is defined as:

\begin{equation}
WF=\frac{{\partial}\tau(v,\theta,p)}{{\partial}lnp}
\label{eq10}
\end{equation}

$\tau(v,\theta,p)$ is the transmittance of the channel with central wavenumber $v$, which depends on the absorption coefficient of absorbed gas in the atmosphere and their vertical distribution of density. $p$ is the pressure and $\theta$ is the satellite zenith angle. Clearly, the WF represents the contributions of the atmosphere at different altitudes to the radiance observed by the satellite.
The Jacobian is used to evaluate the sensitivity of a radiance to a physico-chemical parameter. For a channel with central wavenumber $v$, it represents the sensitivity of the brightness temperature with respect to a change in a geophysical parameter ($X$) such as humidity in our case. It is defined as: 

\begin{equation}
J_v(X)=\frac{{\partial}BT(v)}{{\partial}X}
\label{eq11}
\end{equation}

\begin{figure}[ht]
\centering
\includegraphics[width=1.0\textwidth]{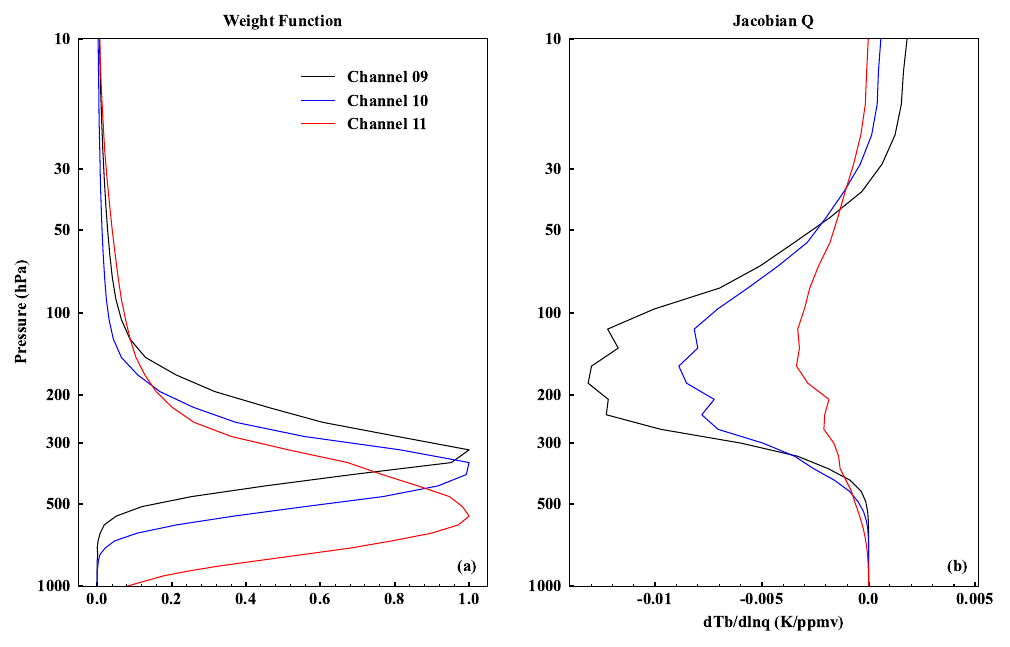}
\caption{Normalized weighting functions (a) and humidity Jacobian functions (b) for channel 9-11. The atmospheric profile comes from the US standard atmosphere. The rapid radiative transfer model adopted is RTTOV version 13.2.}
\label{fig:Jacobian}
\end{figure}

\end{document}